\documentclass[letterpaper, 10 pt, conference]{ieeeconf}

\usepackage{diagbox}
\usepackage{enumerate}
\usepackage{graphicx}
\usepackage{multirow}
\usepackage{url}

\usepackage{CJK}

\usepackage{hyperref}

\IEEEoverridecommandlockouts                              

\title{\LARGE \bf
Room Detection for Topological Maps
}

\author{S\"oren  Schwertfeger$^{1}$ and Tianyan Yu
\thanks{$^{1}$Both authors are with the School of Information Science and Technology, 
	ShanghaiTech University, China.
	{\tt\small <soerensch>@shanghaitech.edu.cn}
}%
}

\begin{document}

\maketitle
\thispagestyle{empty}

\begin{abstract}
Mapping is an important part of many robotic applications. In order to measure the performance of the mapping process we have to measure the quality of its result: the map. The map is essential for robotic algorithms like localization and path planning. Previously it was shown how matched Topology Graphs can be used for map evaluation  by comparing the topology of the robot generated map to the topology of a ground truth map. In this paper we are extending the previous work by detecting open areas, for example rooms, in the 2D grid map and adding those to the topological representation. This way we can avoid the unreliable generation of paths in open areas, thus making the Topology Graph generation, and through that also the Topology Graph matching, more stable and robust. The detection applies the alpha shape algorithm for room detection.\\

\textbf{Keywords:} Robotics; Topological Map; Room Detection; Alpha Shape;

\end{abstract}

\section*{Translation Notice}
This is an English version of this Chinese Journal paper by the same authors:

\begin{CJK}{UTF8}{gbsn}
S{\"o}ren Schwertfeger and 于天彦. 拓扑地图中的房间检测. In {\it 电子设计工程}, 2016. (S\"oren Schwertfeger and Tianyan Yu. Room detection for topological maps. In Electronic Design Engineering). \url{http://mag.ieechina.com/OA/darticle.aspx?type=view&id=201606004}
\end{CJK}

\section{Introduction}

In robotics, a map of the environment can be generated by robots using laser scanners by employing Simultaneous Localization and Mapping (SLAM) techniques \cite{ThrunMapSurvey}. The generated maps are models of the environment which are often represented as 2D grid maps.  Topology Graphs are more abstract representations which only comprise of places and connections between them.

There are many applications for topological maps, for example for map merging \cite{Saeedi-TopologicalMapMerging_RAM2014}, place detection \cite{Beeson2005}, or planning \cite{Thrun-LearningMetroicTopologicalMaps-AI1998}. There are also different ways to generate topological representations from 2D grid maps, for example based on thinning methods \cite{Ko2004} or Voronoi Diagrams \cite{Kolling-surveillanceGraphs-IROS2008}\cite{Lau-VoronoiDiagrams-IROS2010}.

In our previous work we used topological maps to evaluate the quality of the underlying 2D grid maps. This is interesting since all maps carry some degree of error, up to the point of severely bend or broken maps due to localization errors \cite{SSRR11-Schwertfeger-Fiducials}.
Our work determined the map quality by matching topological maps and measuring the error of the match \cite{Schwertfeger2015_Topo_AuRo}\cite{MapEvaluation-TopologicalStructures-ICRA13}. This worked well for relatively confined spaces and hallways. But in more open areas and rooms the underlying Voronoi Diagram is very sensitive to noise on the wall, thus generating different graphs inside open areas for different maps of the same environment. Those different graphs can then not be matched against the ground truth graph.

To alleviate this problem we propose to represent the room as a single vertex in the topology graph with edges connecting to it from all adjacent corridors. For that we first have to detect the room and then cut the topology graph accordingly.

This paper is structured as follows: Section 2 defines a number of terms which are needed in the step-by-step explanation of the generation of Topology Graphs with rooms in Section 3. Section 4 will then present some experimental validation of the algorithm and give a measure of how much better Topology Graphs with rooms can be matched to each other as compared to Topology Graphs without rooms. Section 5 will then present a conclusion of this paper.

\section{Definitions}

 Here are some important definitions for following sections.

  \begin{itemize}

  \item {\textbf{Graph:}}
  A graph is an ordered pair $\mathbf{G=(V,E)}$ comprising a set $\mathbf{V}$ of vertices or nodes together with a set $\mathbf{E}$ of edges or links.
  \item {\textbf{Topology Graph:}} A graph in which vertices represent locations and edges which represent the fact that there exists a drivable route between two vertices.

  \item {\textbf{Vertex:}} A node in the graph. It is attributed with the metric location as x, y coordinates.

  \item {\textbf{Edge:}} A (drivable) connection between vertices in the graph. Each edge is attributed with a metric Path.

   \begin{itemize}

   \item{Half Edge:} In a Doubly Connected Edge List, a half edge is a directed connection between two vertices.
   \item{Twin:} Each half edge has one twin. A half edge's source vertex is its twin's target vertex and vice versa.

   \end{itemize}

  \item {\textbf{Path:}}
  {Every edge is attributed with exactly one path. The path represents the metric information of a free (drivable) way between two vertices. A path is a small directed graph consisting of a series of edges which are connected in a (possibly curved) line between the source vertex and the target vertex of the parent edge. Each vertex in the path is attributed with its metric x, y coordinates.}

  \begin{itemize}

  \item{Half Path:} A half path is a directed path between two vertices.
  \item{Twin:} Each half path has its twin. A half path's source vertex is its twin's target vertex.
  \item{Dead End:} A half path whose target or source vertex only connect to itself and its twin.

  \end{itemize}

 \item {\textbf{Room:}}
  {Generally, rooms are big empty spaces with no obstacles inside, which means that the distance between every two obstacles on its border must be bigger than a certain threshold. In this approach, they are presented by the polygons generated by Alpha Shape with a certain squared value.}

  \begin{itemize}

  \item{Border Vertex:} The vertex at the intersection of a certain path and a certain room.
  \item{Room ID:} The identification number of a room. Default value is $-1$.
  \item{Room Path:} The path connects an inner vertex in a room with another one in the same room (or a dead end's source, or a target vertex with 2 edges connected with the same room ID).
  \item{Room Center:} The centroid vertex of a room which is newly generated after room detection.
  \item{Room Path:} The straight line paths between the room center to all the border vertices of a room which is also newly generated after room detection.

  \end{itemize}

  \end{itemize}

\section{Room Detection for Topology Graphs}

Rooms are big and empty spaces in the environment which are represented with few or no obstacle cells in the corresponding area of the 2D grid map. As can be seen in Figure 2 the Topology Graph edges and paths in the rooms are sensitive to the noise on the wall. In the original approach we used the biggest Alpha Shape polygon to get the outer bounder of the whole map. This polygon was then used to filter away the outside parts of the Voronoi Diagram .

Here we are generating inner polygons with the Alpha Shape algorithm and treat them as distinct rooms. We then cut the path crossing the polygons into two new paths and also generate a new vertex on the intersection between the polygon and the path. Additionally we generate  the centroid vertex of each room. All vertices except the border ones and all inner edges of every room are removed while new paths are set up as straight lines which directly connect the border vertices and the centroid vertex of each room.

\begin{subsection}{Generating the Topology Graph}

Here we will shortly summarize of the Topology Graphs are generated - more details can be found in \cite{Schwertfeger2015_Topo_AuRo}\cite{MapEvaluation-TopologicalStructures-ICRA13}. We are using Voronoi Diagrams to generate a topological graph from a 2D grid map.
The Voronoi Graph is a partition of the space into cells \cite{Klein1988}. All occupied cells in the 2D grid map are used as obstacles in the Voronoi Graph. Each cell of the Voronoi Graph contains all points whose distances to its center are less than those to the other cell's, which generates a skeleton of the map.

  \begin{center}\small
  \includegraphics[width=\linewidth]{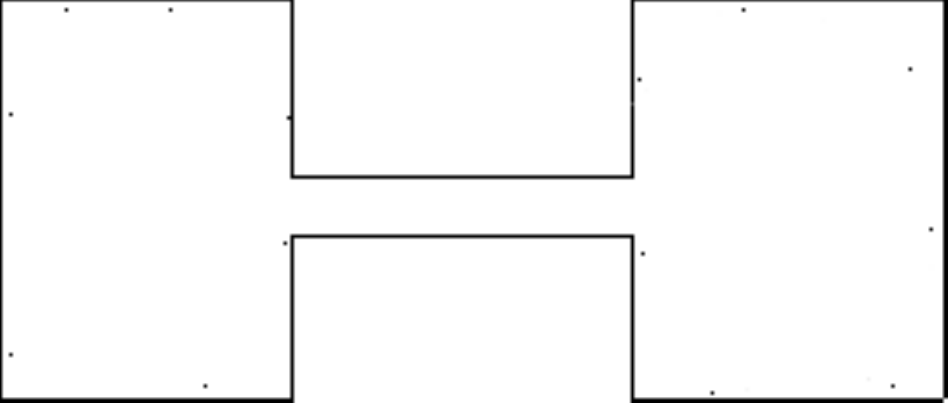}\\
   Fig. 1\quad   Artificial Map A with artificial noise close to the walls.
  \end{center}
  \begin{center}\small
  \includegraphics[width=\linewidth]{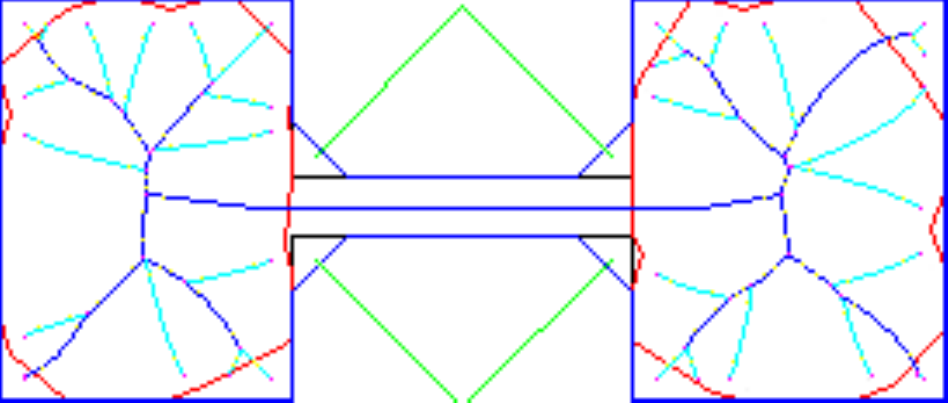}\\
   Fig. 2\quad   The Alpha Shape in Artificial Map A (blue: outside alpha shape; red: inside alpha shape) and the Voronoi Diagram (VD) (green: edges to the outside; blue and turquoise: edges of the VD).
  \end{center}
  \begin{center}\small
  \includegraphics[width=\linewidth]{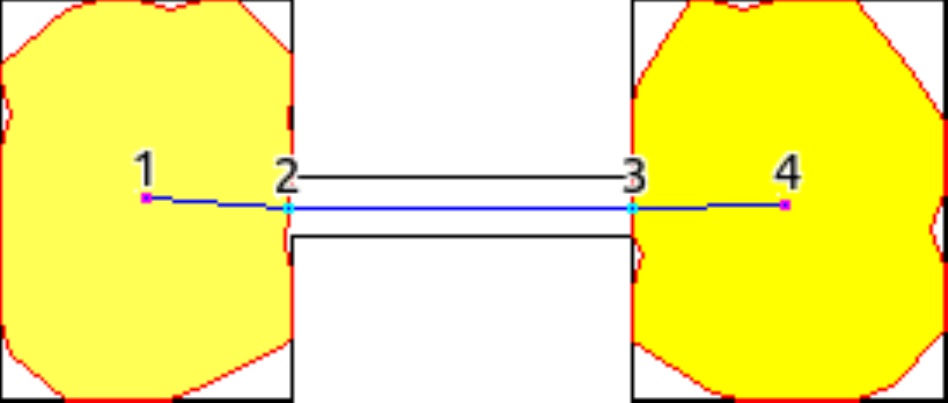}\\
   Fig. 3\quad  Artificial Map A with the Room Vertices (1 and 4); Border Vertices (2 and 3). The edges between the vertices in blue, alpha shape polygons in red and room area as found by the alpha shape in yellow.
  \end{center}

We then filter and prune the Voronoi Graph by filtering out edges too close to an obstacle cell.
     After filtering, the death ends (edges leading to vertices only connected one other vertex) are removed, and the vertices which are very close to each other are merged.

     In the end we have a topological map with vertices representing locations, edges representing a free space connection between the vertices and paths attributed to the edges which are a metric representation of the shape of the path between two vertices.
\end{subsection}

\begin{subsection}{Finding the Rooms}
  In this paper we take the Topology Graph generated above and remove the edges and vertices inside a room. The first step for that is to detect the room. For that we are using the Alpha Shape algorithm.
  Alpha shape is a family of piecewise linear simple curves in the Euclidean plane associated with the shape of a finite set of points. It was first defined by Edelsbrunner, Kirkpatrick and Seidel \cite{Edelsbrunner83}.
 The alpha value determines the minimum distance between obstacle

  \end{subsection}

  \begin{subsection}{Generating the Border Vertices}
    Now that we have the alpha shape polygons for the different rooms we add new vertices called "Border Vertex" at all intersections between the path of the edges and the polygons.
    For that we simply check for each of the path edges/ segments (whose source and target vertex are attributed with (x, y) coordinates) if and where they intersect with the polygon.

    The result is then returned as a list of border vertex locations, defined as the distances from the source vertex of that edge to the intersection point. This distance will then be used during the cutting procedure in the next step.

  \end{subsection}

  \begin{subsection}{Cutting a Path at a Border Vertex}
  For all found border vertices the cut distance from the source vertex of its edge to the border vertex is between $0$ and the length of that path. The result of this step is a list of newly created vertices and their corresponding new edges and associated paths.


There is a special case when the cut distance is close to $0$ or close to the path length (close to the target). Because the distance is represented as an inaccurate floating-point number we use a threshold $\epsilon$ to judge whether the cutting position is close to the source or target vertex of the edge ($\epsilon$ is set to $0.00000001$):

    {$|l_{voric}| \leq \epsilon$ or $|l_{vori} - l_{voric}| \leq \epsilon$}: Cut at the source or target. The list with only $E_{vori}$ is returned.

    The  normal cutting for {$\epsilon < l_{voric} < l_{vori} -\epsilon$} is described next.
    Each halfedge of the topology graph is attributed with a path. This path consists of a list of end-to-end-connected path halfedges (marked as $E_{topo_1},E_{topo_2}, \dots,E_{topo_N}$) with their respective lengths (marked as $l_{topo_1},l_{topo_2}, \dots,l_{topo_N}$). We add those together added one by one. Once the sum of lengths is bigger than the cut distance calculated in the previous step we have found the half edge of the path graph that we have to cut. For that we call the cutting function.

    There are many halfedges ($E_{vori_1},E_{vori_2}, \dots,E_{vori_M}$) in the graph. When the halfedge cutting function is called on a certain halfedge ($E_{vori_k}, 1<k<M$), the returned halfedge list's size will be checked.
    If the size of list is:

    \begin{enumerate}[A)]

    \item{1:} $E_{vori_k}$ is not removed.

        \begin{enumerate}[a)]

        \item {$E_{vori_k}$ is cut at its source}

              $E_{vori_k}$'s source vertex is marked as border vertex.
        \item {$E_{vori_k}$ is cut at its target}

              $E_{vori_k}$'s target vertex is marked as border vertex.

        \end{enumerate}

    \item{2:} $E_{vori_k}$ is removed.

              The same halfedge cutting function will be called on the twin of $E_{vori_k}$.
              Then, a new vertex is set up as a border vertex by adding the newly created paths and their twins which are connected with it to its edge connected list, and it is pushed into the vertices map of the whole graph. The vertex is marked as a border vertex. Its room id is chosen to be the same as the id of the polygon which intersects with the path at the position of the border vertex in the polygon list of the Alpha Shape.


    \end{enumerate}
  \end{subsection}

  \begin{subsection}{Setting Room ids}

    Till now, the newly generated border vertices have their room ids, while the other non-border ones still have default value $-1$ as their room ids. Some of them are in the inner part of a certain polygon, while the others are out of any polygon generated.
    A certain room id, which is chosen to be as same as the polygon's id in the polygon list of the Alpha Shape, should also be given to every non-border vertex inside a certain polygon.

    The room id for the non-border vertices is set by checking every path to find whether its source or target vertex is in a certain polygon in the polygon list. If it's in a certain polygon, the polygon's id in the polygon list in the Alpha Shape is given to the vertex.
    After that, the vertices with the default room id value $-1$ can be regarded as not in any polygons.
    Then, the room id of every path is also set by checking whether the source's room id equals to the target's. If the result is:

    \begin{enumerate}[A)]

    \item True.

        \begin{enumerate}[a)]
        \item Source's room id is $-1$.

        $\Longrightarrow$ Ignore the path.
        \item Source's room id is NOT $-1$.

        $\Longrightarrow$ Set the path and its twin's room id as same as the source's.
        \end{enumerate}

    \item False.

        \begin{enumerate}[a)]
        \item If the path is a death end and not longer than a threshold (\textbf{minLenThreshold}):

        As the room id of the path's end vertex (the source or target vertex with only 2 paths connected) must be the default value $-1$, the room id of another vertex with non-default room id is given to the path as its room id, which will be useful in the dead end removal procedure.

        \end{enumerate}
    \end{enumerate}

    In the experiments below the threshold minLenThreshold is set to $20$ pixels.
    \end{subsection}

    \begin{subsection}{Removing Short Dead Ends}

	Dead ends are edges that connect to a vertex which is only connected to two halfedges: this edge and its twin. The according vertex is then also labelled as dead end. If a border vertex is connected to a dead end edge whose length is less then the minLenThreshold, all of them are removed from the graph: The dead end vertex, the dead end edge, the according border vertex and the edge from inside the room that is connecting that border vertex are going to be removed from the Topology Graph.
	
	We do this because the Alpha Shape does not extend into corners. Thus there might be vertices which are outside of the alpha shape polygon but actually inside the room (for example Fig. 2). Those should be removed since later-on we want to represent the room by a single vertex.

  \end{subsection}

  \begin{subsection}{Creating Topology Graphs with Rooms}

The overall goal of this paper is to create Topology Graphs that represent a room as a single vertex that connects to all "corridors" leading to that room. We have done all the preparation now (found the room, created border vertices with ids and removed short dead ends). So the next step is to remove all edges and vertices (except the border vertices) inside the room.

Furthermore we create a new vertex that represents the room. This vertex is connected by edges to all border vertices of that room and attributed with the polygon describing the room (yellow area in the figures). It is also attributed with a location of the centroid of the polygon.

   The centroid of a non-self-intersecting closed polygon defined by $n$ vertices:
   \begin{equation}
   (x_0,y_0), (x_1,y_1), \dots, (x_{n-1},y_{n-1})
   \end{equation}

   is the point $(C_x, C_y)$, where
   \begin{eqnarray}
   &&C_{\mathrm {x} }={\frac {1}{6A}}\sum _{i=0}^{n-1}(x_{i}+x_{i+1})(x_{i}\ y_{i+1}-x_{i+1}\ y_{i})\\
   &&C_{\mathrm {y} }={\frac {1}{6A}}\sum _{i=0}^{n-1}(y_{i}+y_{i+1})(x_{i}\ y_{i+1}-x_{i+1}\ y_{i})
   \end{eqnarray}
   and where $A$ is the polygon's signed area,
   \begin{eqnarray}
   &&A={\frac {1}{2}}\sum _{i=0}^{n-1}(x_{i}\ y_{i+1}-x_{i+1}\ y_{i})
   \end{eqnarray}

   In these formulas \cite{BourkeWebPolygonArea1988}, the vertices are assumed to be numbered in order of their occurrence along the polygon's perimeter. Furthermore, the vertex $(x_n,y_n)$ is assumed to be the same as $(x_0,y_0)$, meaning $i+1$ on the last case must loop around to $i = 0$.

 \end{subsection}

\def \myGraphicsWidth {0.95}
  \begin{center}\small
  \includegraphics[width=\myGraphicsWidth\linewidth]{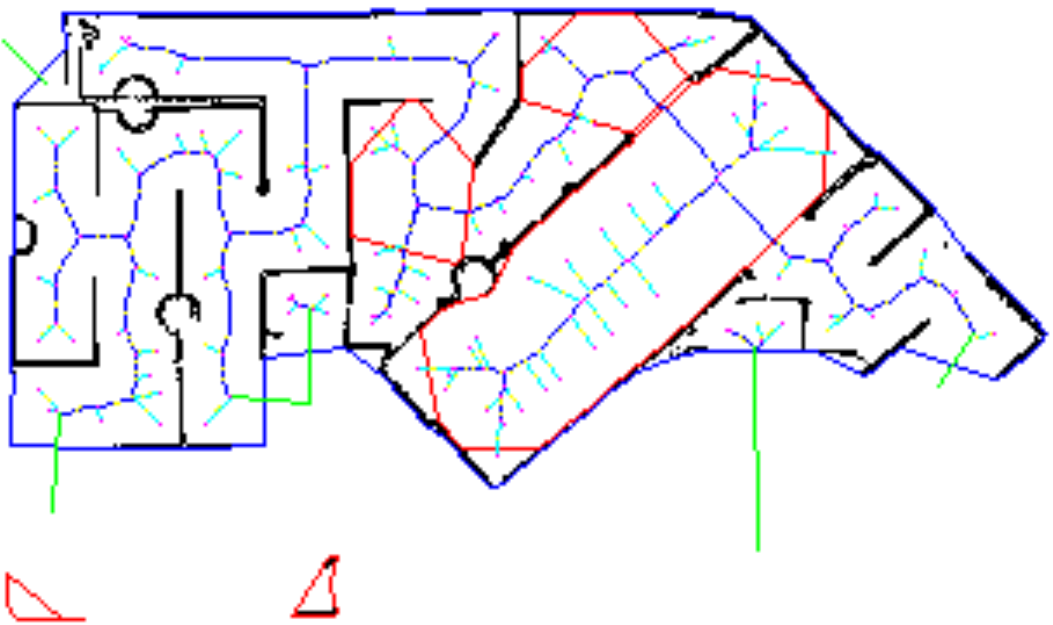}\\
   Fig. 4\quad  RoboCup Rescue map with VD and Alpha Shape
  \end{center}

  \begin{center}\small
  \includegraphics[width=\myGraphicsWidth\linewidth]{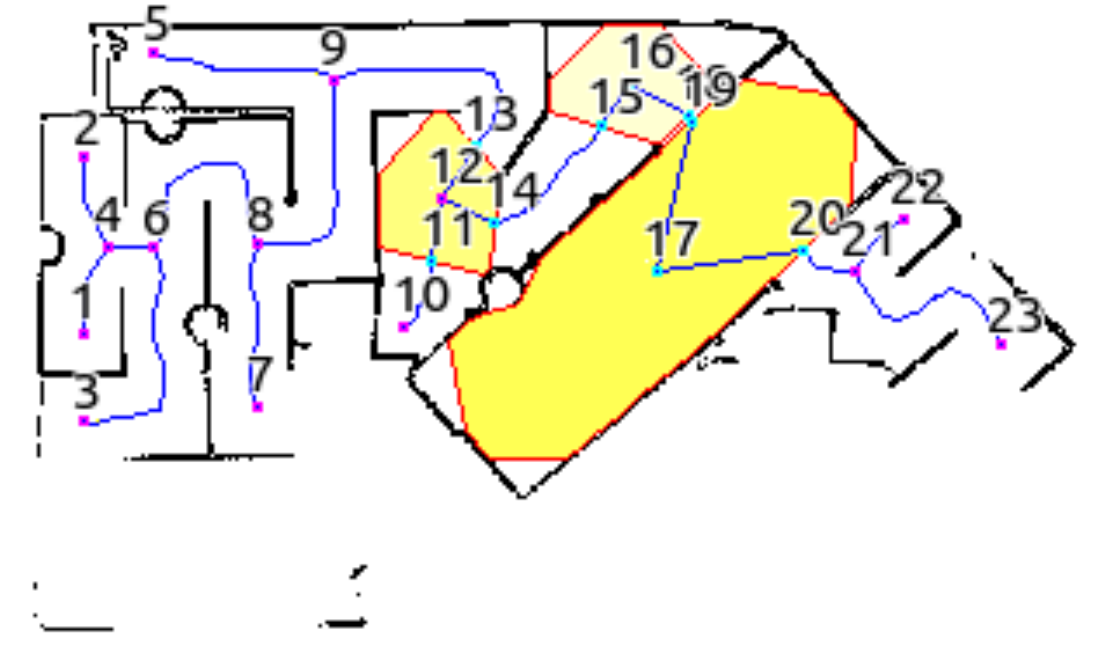}\\
   Fig. 5\quad  Topology Graph of Fig. 4 with numbered vertices and three detected rooms in yellow
  \end{center}

  \begin{center}\small
  \includegraphics[width=\myGraphicsWidth\linewidth]{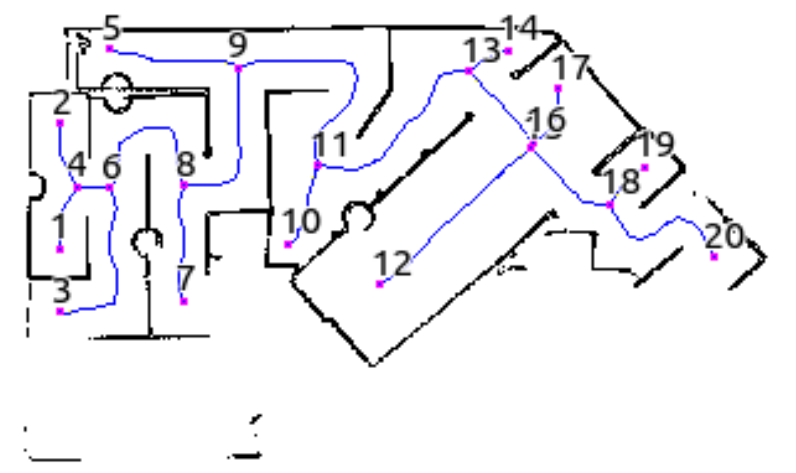}\\
   Fig. 6\quad  Topology Graph of Fig. 4 without room detection
  \end{center}

\section{Experiments}
 First we show that the algorithm generally works and then we make extensive experiments to measure the improvement that room detection offers for graph matching.

   \begin{center}\small
  \includegraphics[width=\myGraphicsWidth\linewidth]{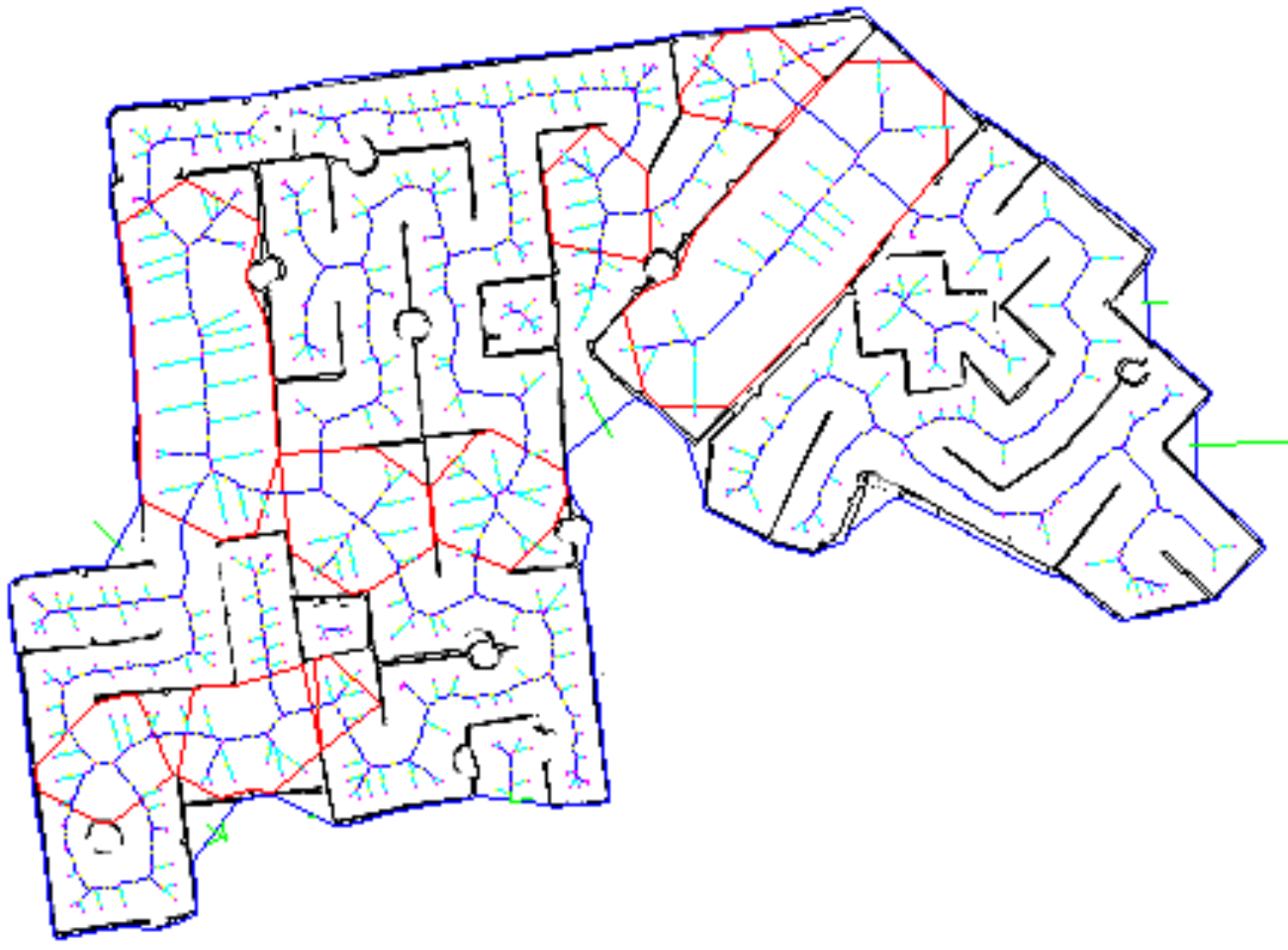}\\
   Fig. 7\quad  RoboCup Rescue map with VD and Alpha Shape
  \end{center}

  \begin{center}\small
  \includegraphics[width=\myGraphicsWidth\linewidth]{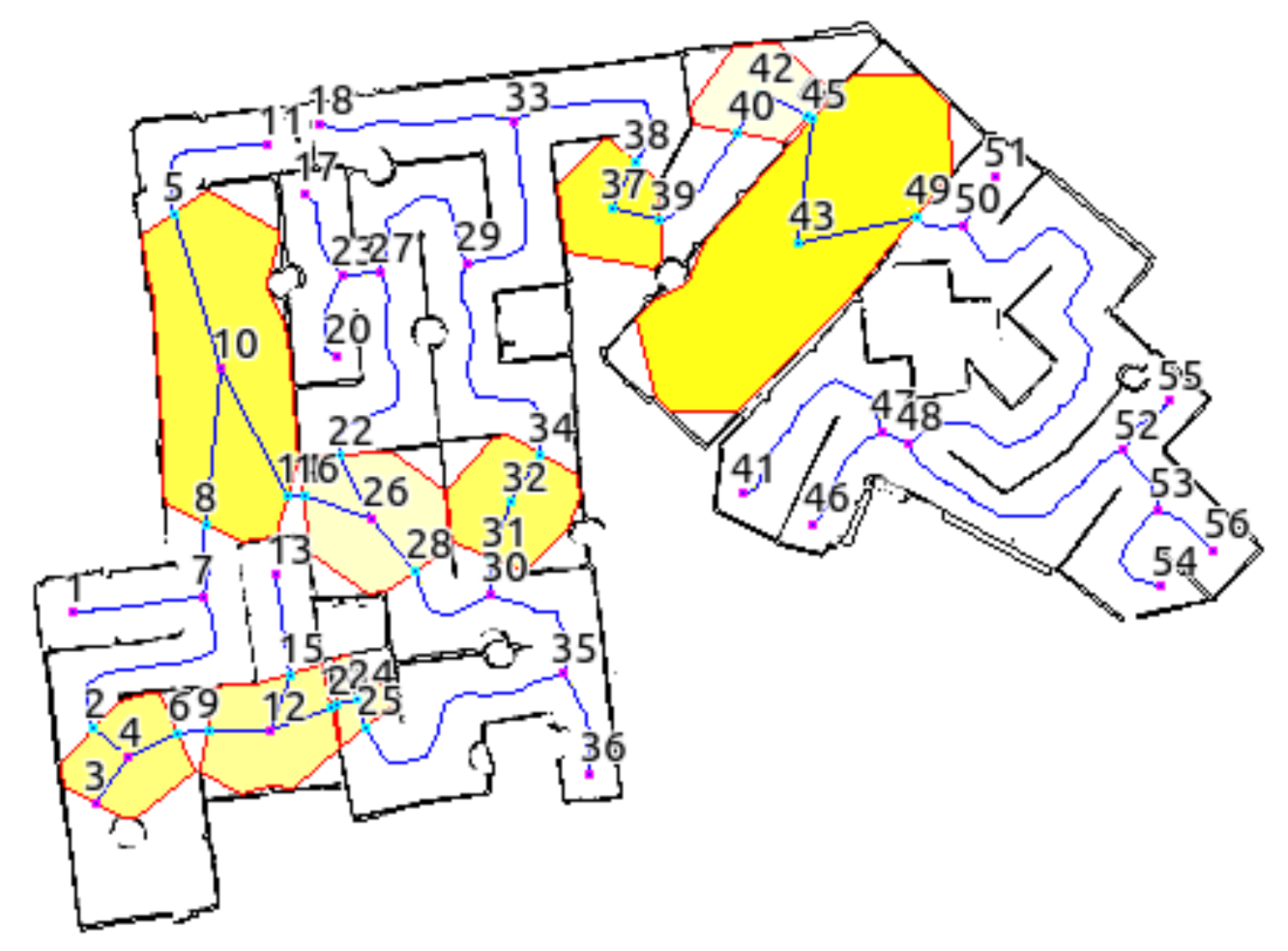}\\
   Fig. 8\quad  Topology Graph of Fig. 7 with numbered vertices and three detected rooms in yellow
  \end{center}

  \begin{center}\small
  \includegraphics[width=\myGraphicsWidth\linewidth]{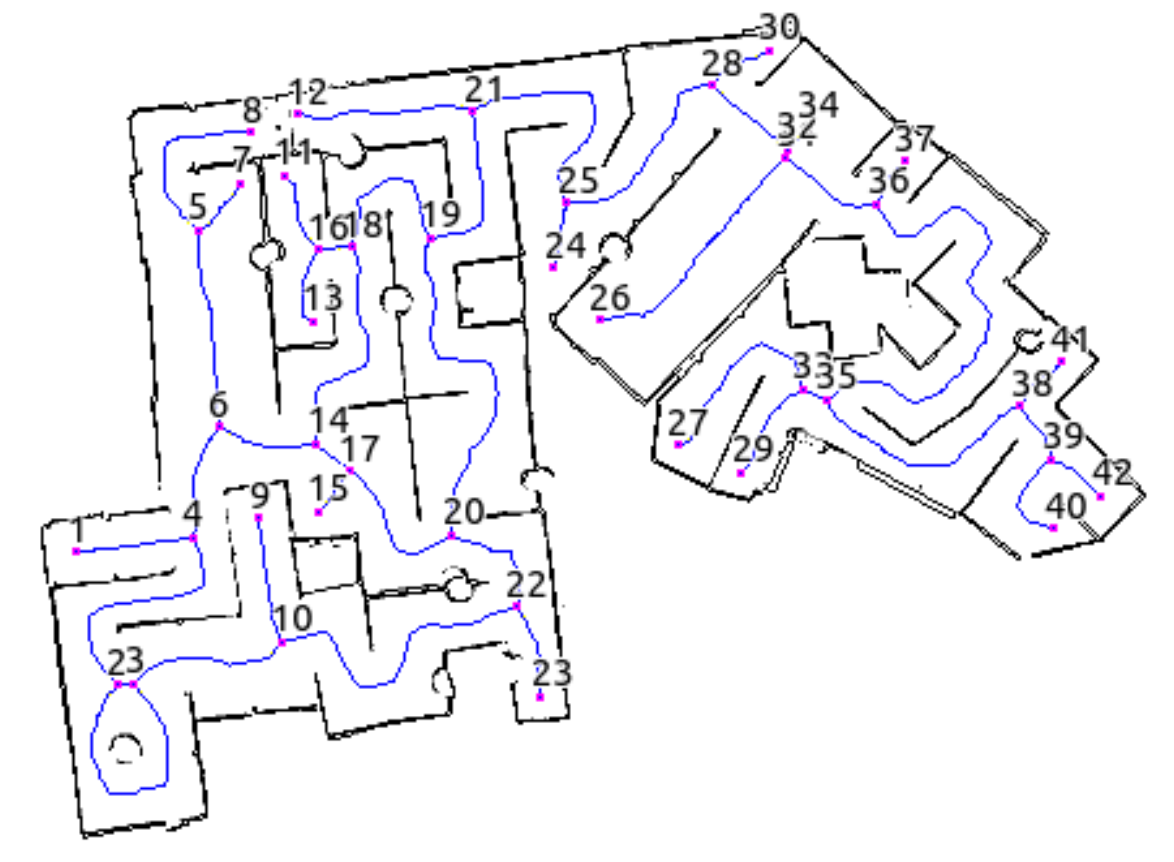}\\
   Fig. 9\quad  Topology Graph of Fig. 7 without room detection
   \end{center}

   \begin{subsection}{Artificial map}
  Fig,~1-3 shows an artificially created map of two rooms connected by a corridor. Some noise was added to the walls in form of a few occupied cells. In Figure 2 one can see that due to this noise the Voronoi Diagram (VD) of the two rooms differ significantly, even though the rooms are (except the noise and the fact that one is mirrored) the same. This shows why generating VDs inside open areas will lead to more or less randomly placed vertices, which in turn are hard to match between two maps. Figure 2 also shows the Alpha Shape polygons that will be used for the room detection. Figure 3 then shows the vastly simplified Topology Graph with just two Room Vertices (1 and 4) and two Border Vertices (2 and 3).

  \end{subsection}

    \begin{subsection}{RoboCup Rescue Maps}

A more realistic example is given using maps from RoboCup Rescue \cite{sheh2011robocuprescue} \cite{sheh2012advancing}. Those maps were also used in previous experiments with the Topology Graphs and their 3D versions also for 3D map evaluation\cite{Schwertfeger-3DMapEval-SSRR15}. Figures 4 and 7 show maps and the complexity of the VD and the alpha shapes while Figures 5 and 8 show the filtered Topology Graph with the detected rooms. Compared to the Topology Graphs without room detection (Figures 6 and 9) we can see the reduced complexity of the graphs.

  \end{subsection}

  \begin{subsection}{Effect on Matching}
     In order to show the room detection's role in generating more reliable and consistent Topology Graphs from a 2D grid map, a map (we call it "Beeson map" here) from the paper \cite{Beeson2005}, a map ("Belgioioso") from \cite{HaehnelBelgioiosoDataset}
     and three more maps from Radish \cite{Radish} (" Edmonton Convention Center", "Oakmont", "SDR Site B") are chosen.

     We pre-process the data in two steps:

     \textbf{Step 1:} Threshold for black: Threshold the color values of the occupied cells for the map. If the sum of a pixel's RGB value ($R+G+B$) is not greater than the threshold, it will be set to black (occupied), otherwise it will be set to white (free).

     \textbf{Step 2:} Thinning: Remove all the black pixels (occupied) which are surrounded by other occupied neighbour pixels (in 4 or 8 directions) in the map. This way we remove occupied pixels inside solid areas. This is done just to save computation time when generating the Vorodnoi Diagram and has no effect on the Topology Graph.

     For the following experiments the threshold for black is set to 100, and the thinning is in 8 directions. Figures 10 - 13 show the maps and the Topology Graphs with the detected rooms.

     \begin{center}\small
     \includegraphics[width=\linewidth]{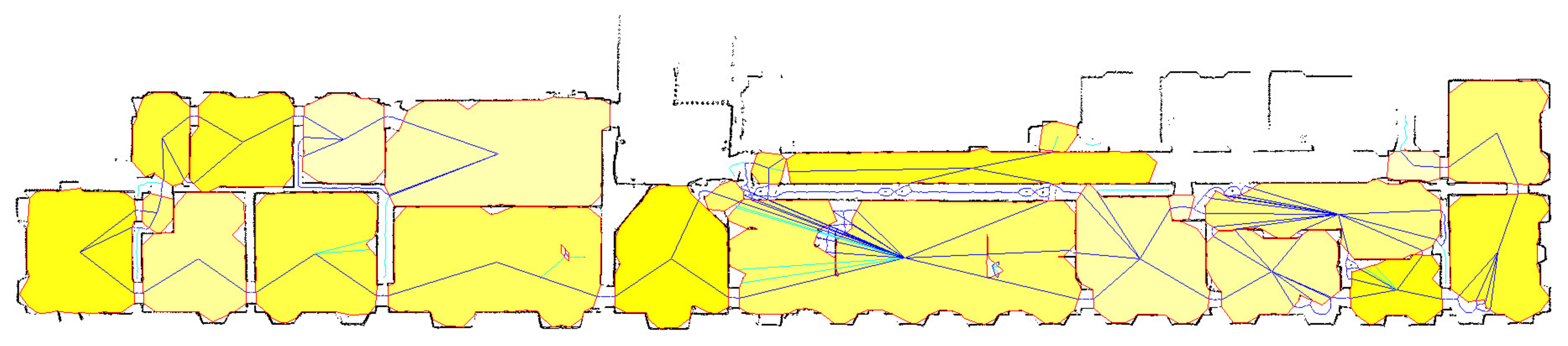}\\
     Fig. 10\quad  Belgioioso map
     \end{center}

     \begin{center}\small
     \includegraphics[width=\linewidth]{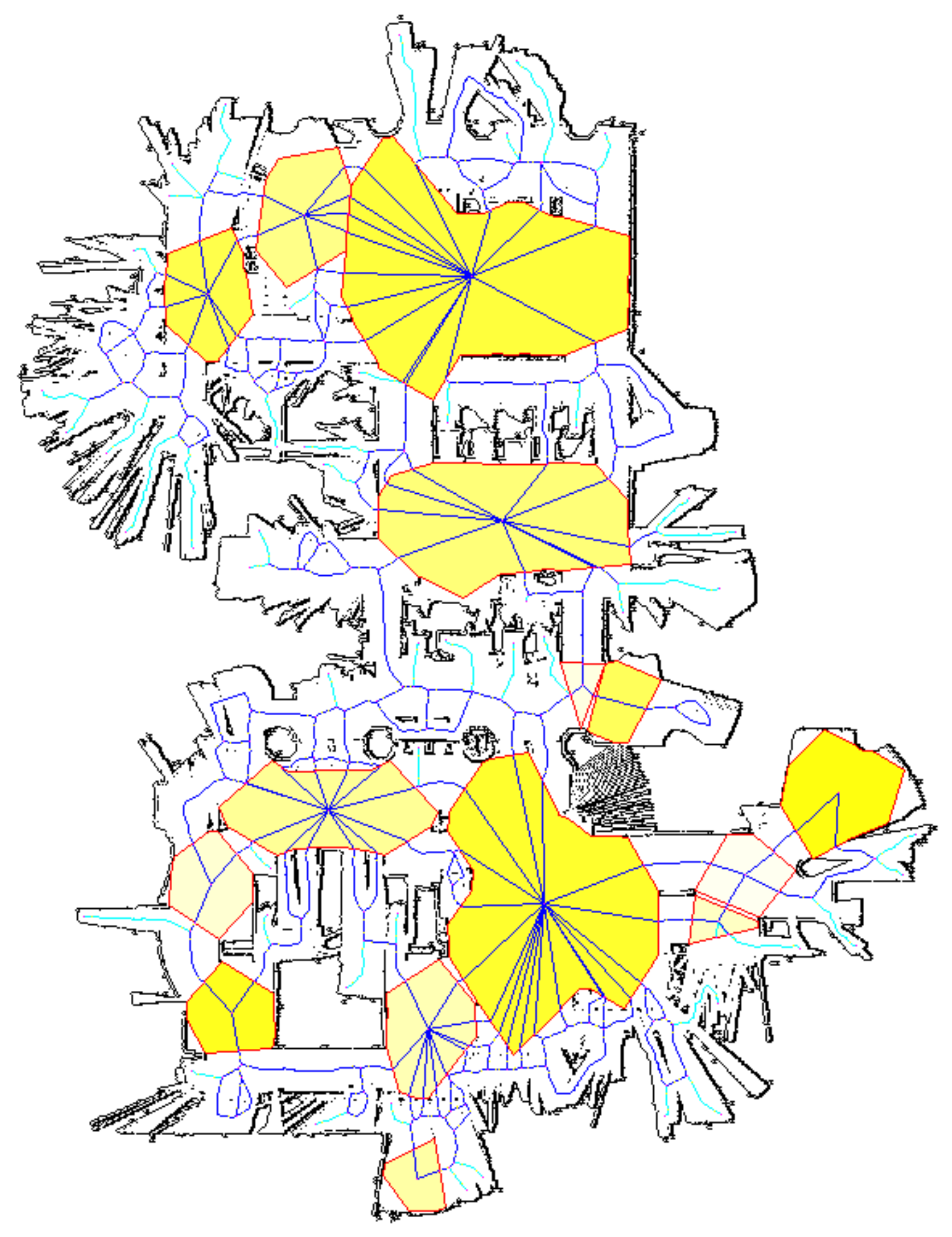}\\
     Fig. 11\quad  Edmonton Convention Center
     \end{center}

     \begin{center}\small
     \includegraphics[width=\linewidth]{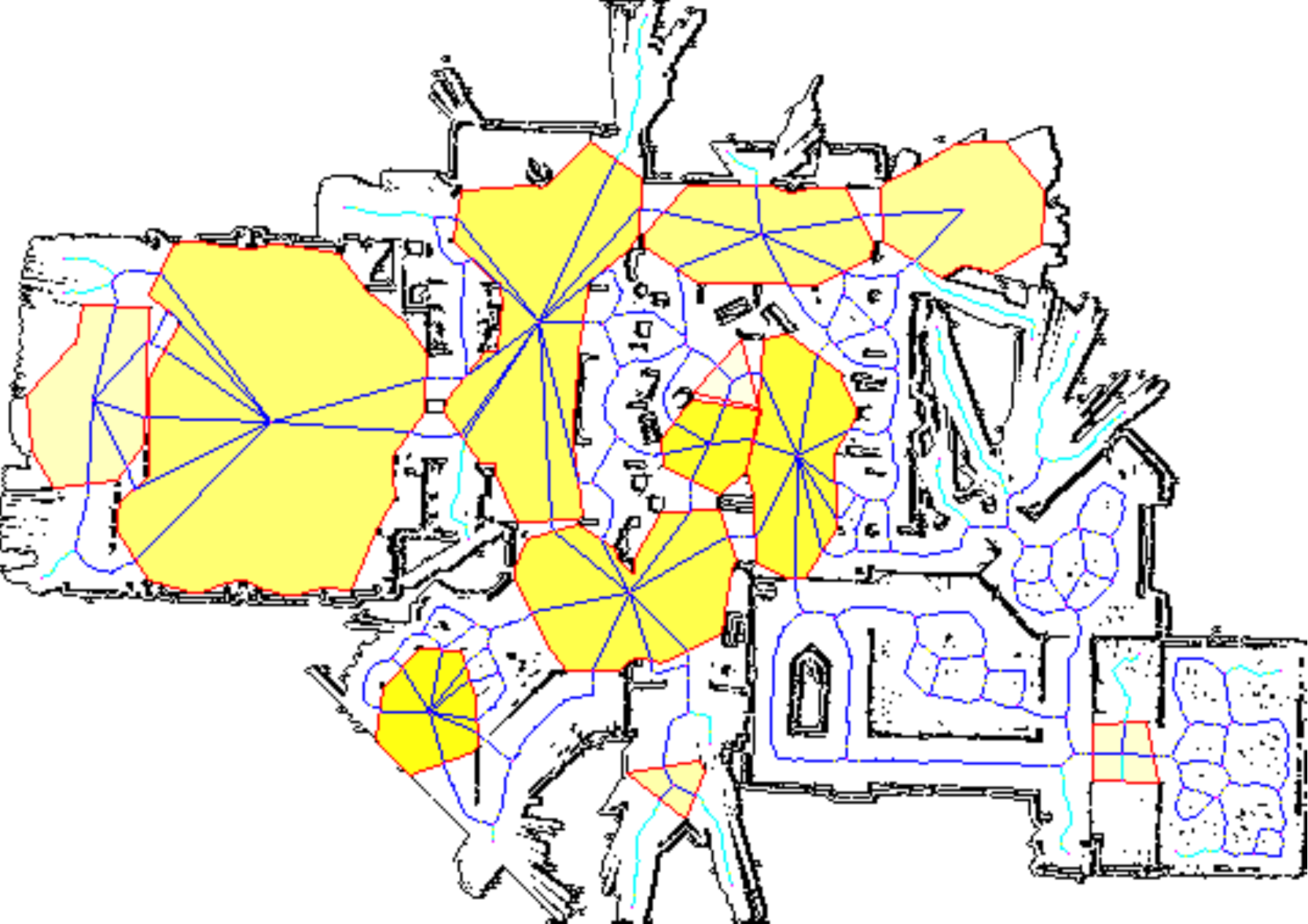}\\
     Fig. 12\quad  The Longwood at Oakmont Nursing Home
     \end{center}

     \begin{center}\small
     \includegraphics[width=\linewidth]{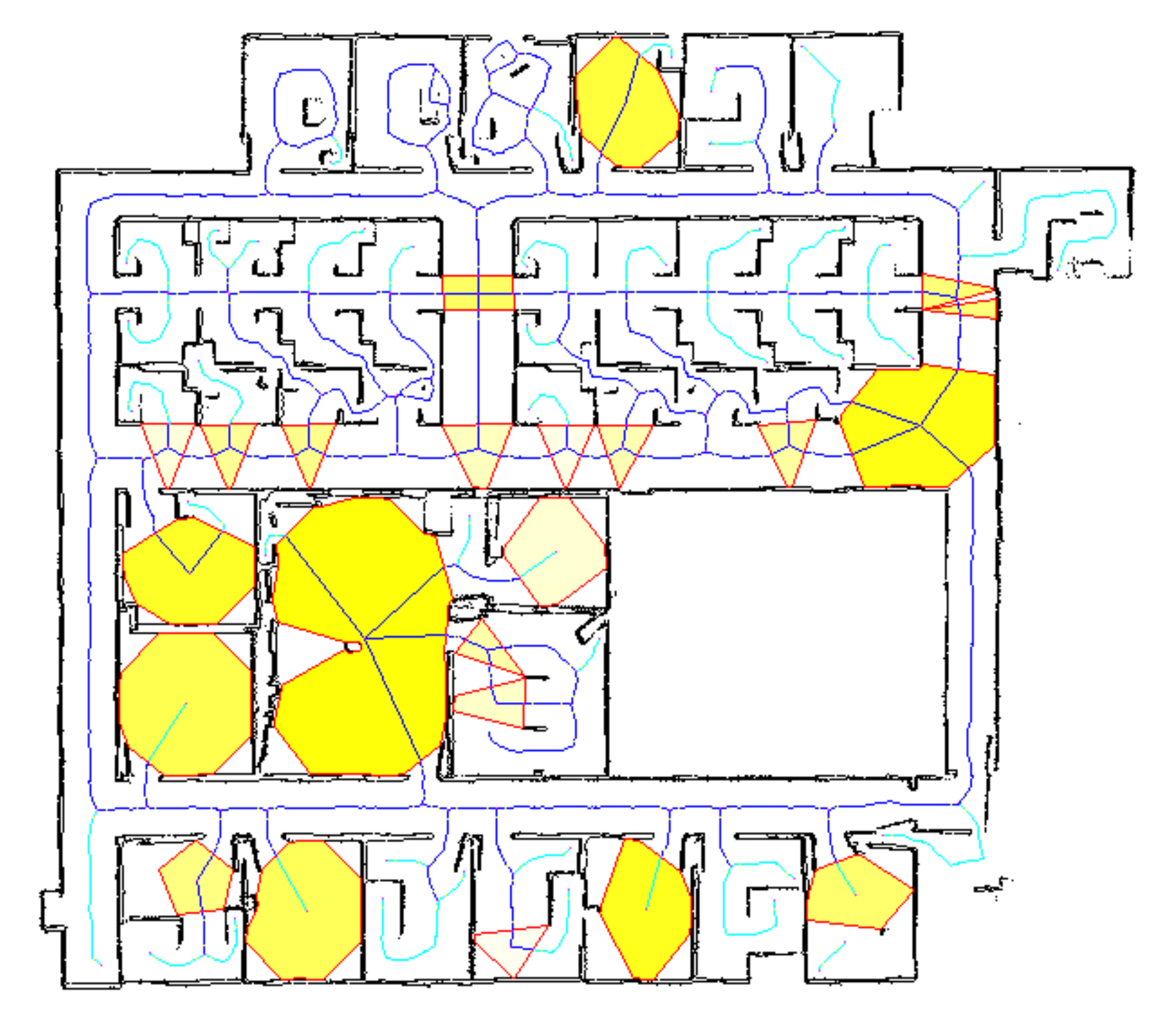}\\
     Fig. 13\quad  The SDR Site B
     \end{center}

     In the next step we create several versions of each map with random noise near the walls. We do this by removing a percentage of black pixels and adding the same amount of new random pixels with a distance of less than a certain value $d_r$ away in each axis from existing, randomly chosen occupied pixels. With this we simulate multiple mapping runs with different amount of noise in the maps.

    During the following experiments, $d_r$ is chosen from $\{1,2,3,5\}$, and the percentage of removing ($p$) is chosen from $\{2\%,5\%,8\%,11\%,14\%\}$.
     Figure 14 shows the Beeson map with maximum random distance of 5 pixel and several percentages of randomized pixel. The room parts are filled with different shades of yellow to better visualize the results.

     For each pair $\{d_r,p\}$, $10$ randomization maps are generated for the test. One map ( $d_r = 5$; $p = 2\%$) is also generated to serve as the base map for matching to all other maps.
     We then match the Topology Graph of the base map to each of the other map's Topology Graphs. We simply do this matching by just matching the vertices.
     For each vertex of the base map we search for the closest (geometric distance) vertex in the other, randomized map. We sum up and then average the geometric distances of all such matches as a measure of the stability, reliability and repeatability of the generated Topology Graphs. For each randomization configuration we average the result of the ten generated maps. We do those experiments for Topology Graph generation with and without room detection.

          \def \myRandomMapWidth {0.89}

     \begin{center}\small
     \includegraphics[width=\myRandomMapWidth\linewidth]{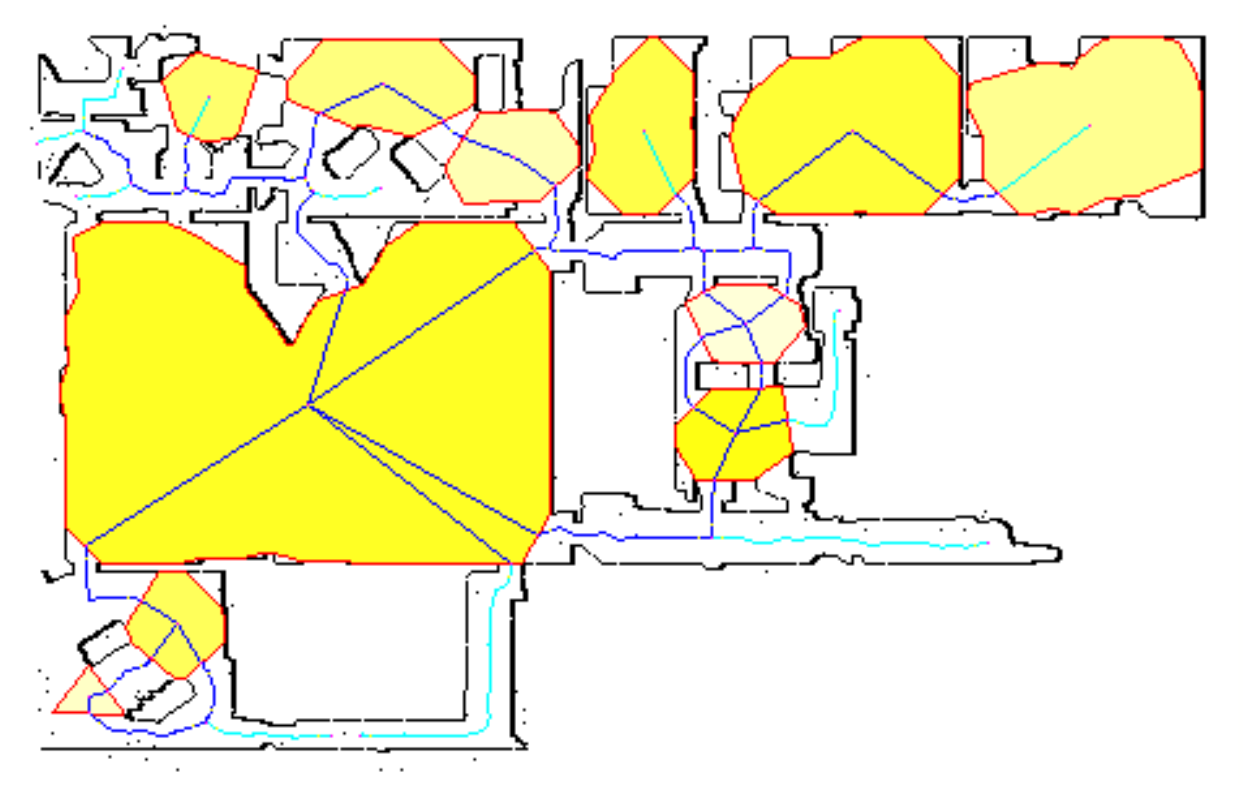}
     \includegraphics[width=\myRandomMapWidth\linewidth]{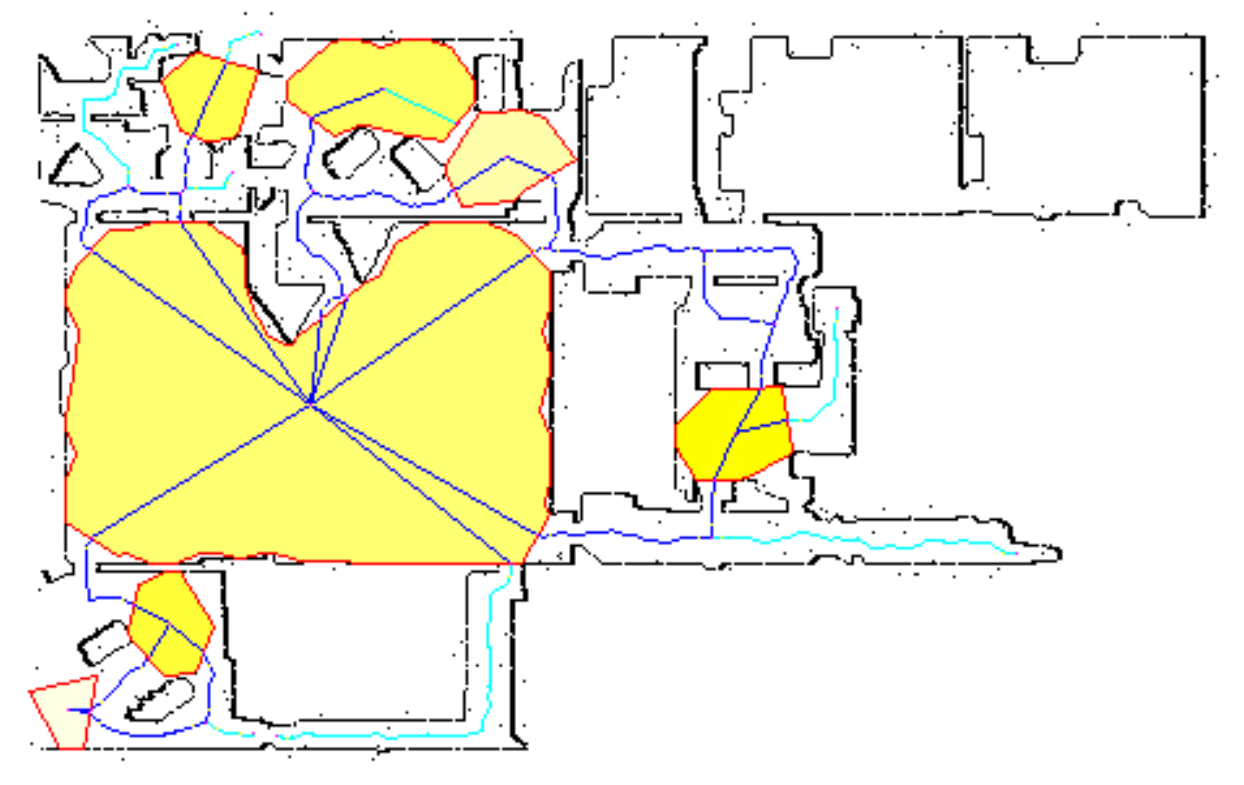}
     \includegraphics[width=\myRandomMapWidth\linewidth]{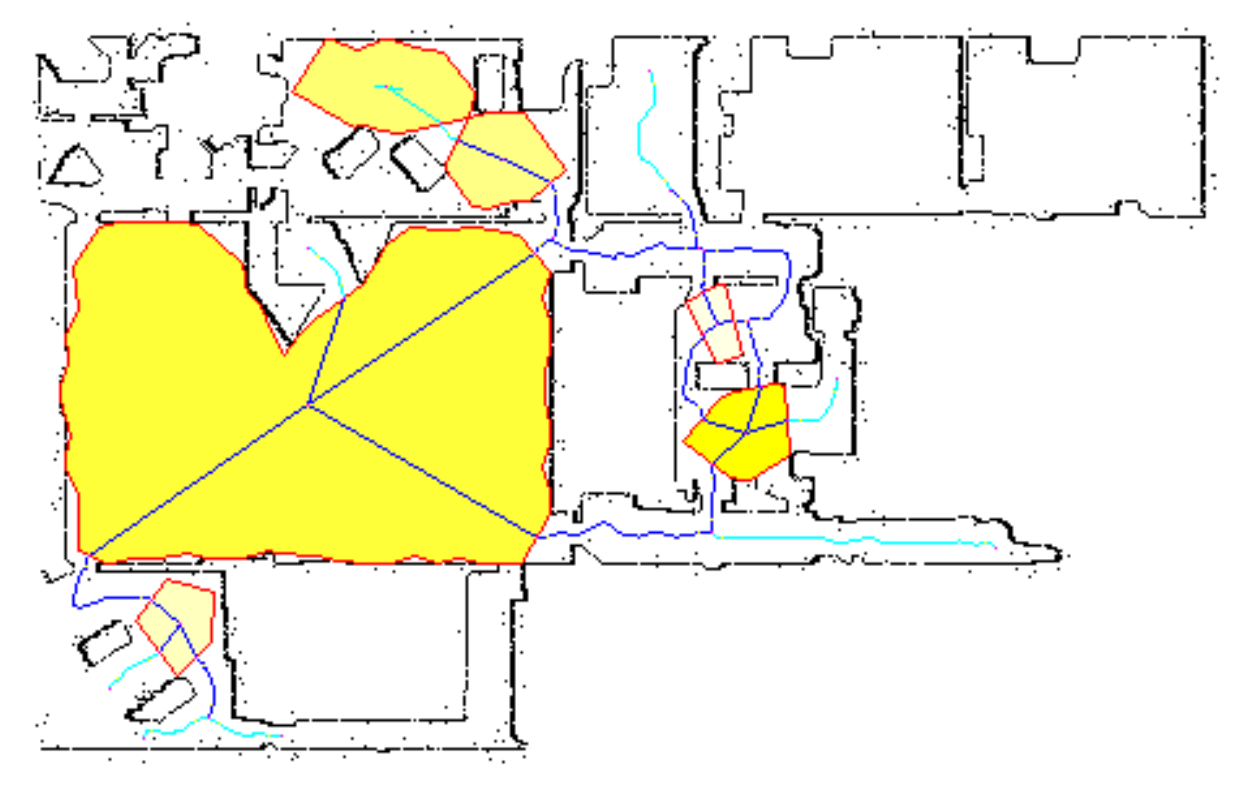}

     \includegraphics[width=\myRandomMapWidth\linewidth]{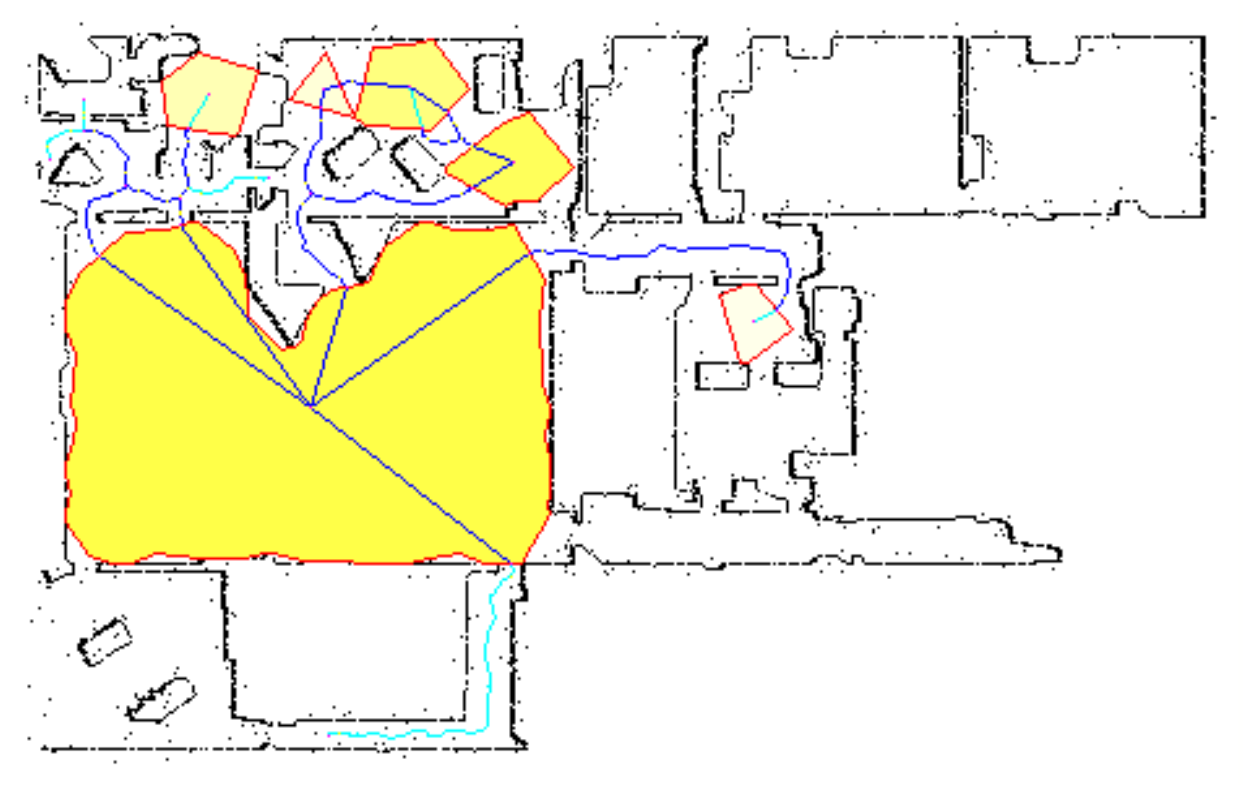}
     \includegraphics[width=\myRandomMapWidth\linewidth]{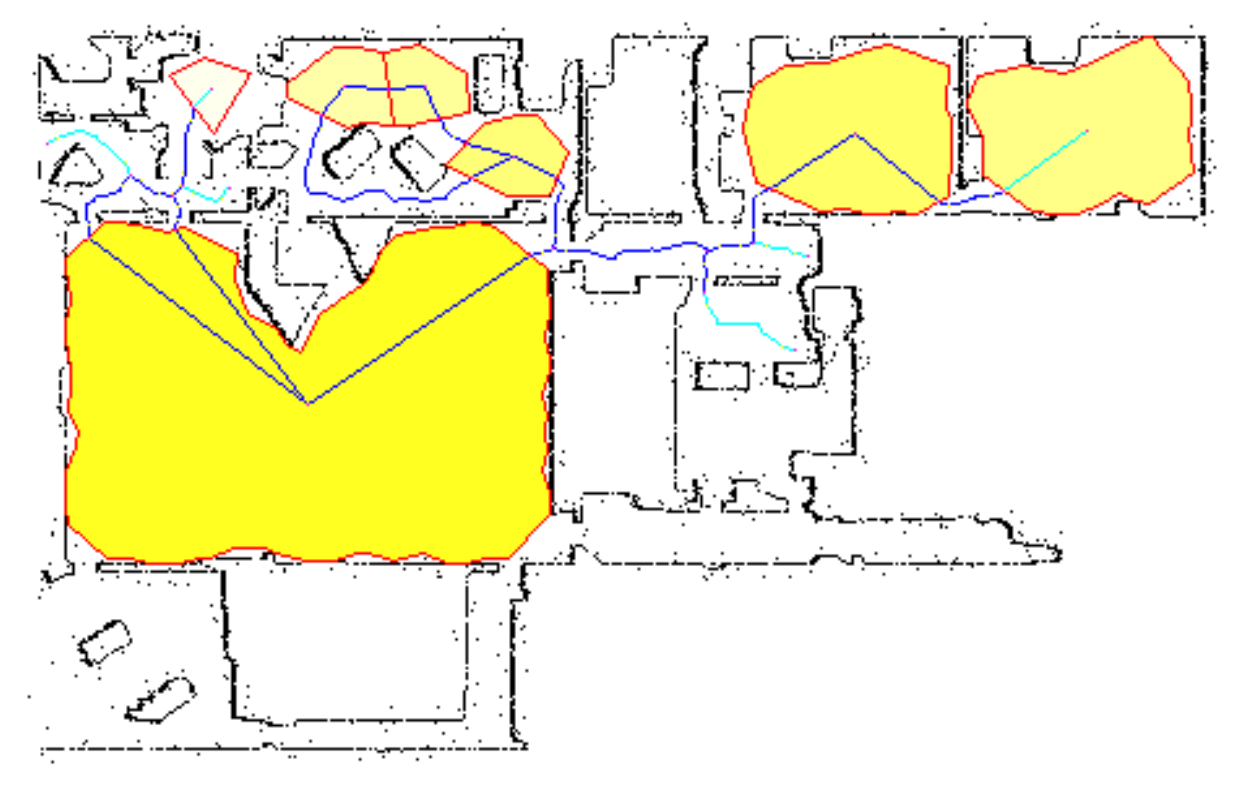}\\
     Fig. 14\quad Room detection on Beeson map randomized (alpha value = 200; $d_r = 5$; $p = 2\%$(top)$, 5\%, 8\%, 11\%, 14\%$(bottom) )
     \end{center}

\renewcommand{\arraystretch}{0.95}
\tabcolsep=0.10cm

\begin{center}\small
 {\interlinepenalty=10000
Table 1\quad  Beeson map distance comparison (alpha value: 200)
\begin{tabular}{|cc|cc|c|}
\hline
\multicolumn{2}{|c|}{Parameters}& \multicolumn{2}{c|}{Average Distances} & Ratio\\
Distance & \multicolumn{1}{c|}{Percent} & Room & \multicolumn{1}{c|}{No Room} & \%\\

\hline
\multirow{5}*{1}
& 2 & 0.91 & 3.16 & 29\\
& 5 & 1.12 & 4.43 & 25\\
& 8 & 1.17 & 4.05 & 29\\
& 11 & 1.52 & 7.10 & 21\\
& 14 & 1.64 & 4.94 & 33\\
\hline
\multirow{5}*{2}
& 2 & 0.59 & 3.88 & 15\\
& 5 & 2.34 & 4.64 & 50\\
& 8 & 6.90 & 10.48 & 66\\
& 11 & 3.98 & 9.24 & 43\\
& 14 & 4.85 & 7.28 & 67\\
\hline
\multirow{5}*{3}
& 2 & 3.73 & 5.61 & 66\\
& 5 & 8.61 & 17.55 & 49\\
& 8 & 6.86 & 12.84 & 53\\
& 11 & 18.97 & 25.71 & 74\\
& 14 & 13.87 & 9.20 & 151\\
\hline
\multirow{5}*{4}
& 2 & 3.94 & 7.60 & 52\\
& 5 & 8.20 & 8.50 & 96\\
& 8 & 15.07 & 13.50 & 112\\
& 11 & 10.82 & 10.82 & 100\\
& 14 & 41.20 & 50.13 & 82\\
\hline
\end{tabular}
}
\end{center}

\begin{center}\small

Table 2\quad  Belgioioso map distance comparison (alpha value: 250)
\begin{tabular}{|cc|cc|c|}
\hline
\multicolumn{2}{|c|}{Parameters}& \multicolumn{2}{c|}{Average Distances} & Ratio\\
Distance & \multicolumn{1}{c|}{Percent} & Room & \multicolumn{1}{c|}{No Room} & \%\\

\hline
\multirow{5}*{1}
& 2 & 0.92 & 2.76 & 33\\
& 5 & 2.66 & 4.67 & 57\\
& 8 & 0.75 & 2.44 & 31\\
& 11 & 3.17 & 6.25 & 51\\
& 14 & 0.94 & 2.57 & 37\\
\hline
\multirow{5}*{2}
& 2 & 0.64 & 0.84 & 76\\
& 5 & 1.85 & 3.78 & 49\\
& 8 & 0.33 & 0.75 & 43\\
& 11 & 2.14 & 4.81 & 45\\
& 14 & 2.72 & 4.64 & 59\\
\hline
\multirow{5}*{3}
& 2 & 2.02 & 3.24 & 62\\
& 5 & 2.36 & 3.32 & 71\\
& 8 & 3.42 & 5.52 & 62\\
& 11 & 2.22 & 3.97 & 56\\
& 14 & 3.50 & 4.62 & 76\\
\hline
\multirow{5}*{4}
& 2 & 1.37 & 3.09 & 44\\
& 5 & 2.74 & 3.02 & 91\\
& 8 & 4.41 & 6.91 & 64\\
& 11 & 2.43 & 2.38 & 102\\
& 14 & 6.62 & 7.13 & 93\\
\hline
\end{tabular}
\end{center}


Tables 1 to 5 show the results of the different maps, respectively. The "Ratio" column shows the
ratio between the averaged distance of the matches of the graphs with room detection and the average distance of the matches of the graphs without room detection.

\begin{center}\small

 {\interlinepenalty=10000
Table 3\quad  Edmonton distance comparison (alpha value: 500)
\begin{tabular}{|cc|cc|c|}
\hline
\multicolumn{2}{|c|}{Parameters}& \multicolumn{2}{c|}{Average Distances} & Ratio\\
Distance & \multicolumn{1}{c|}{Percent} & Room & \multicolumn{1}{c|}{No Room} & \%\\

\hline
\multirow{5}*{1}
& 2 & 0.82 & 1.14 & 72\\
& 5 & 2.02 & 2.33 & 87\\
& 8 & 2.06 & 2.91 & 71\\
& 11 & 2.51 & 3.24 & 77\\
& 14 & 2.87 & 3.47 & 83\\
\hline
\multirow{5}*{2}
& 2 & 1.22 & 1.31 & 93\\
& 5 & 2.07 & 2.65 & 78\\
& 8 & 3.26 & 3.48 & 94\\
& 11 & 3.30 & 4.12 & 80\\
& 14 & 3.42 & 4.17 & 82\\
\hline
\multirow{5}*{3}
& 2 & 1.40 & 1.70 & 82\\
& 5 & 2.65 & 3.13 & 85\\
& 8 & 3.53 & 4.13 & 86\\
& 11 & 4.47 & 4.75 & 94\\
& 14 & 4.63 & 4.73 & 98\\
\hline
\multirow{5}*{4}
& 2 & 2.61 & 2.92 & 89\\
& 5 & 4.06 & 4.57 & 89\\
& 8 & 5.33 & 5.71 & 93\\
& 11 & 5.49 & 5.78 & 95\\
& 14 & 6.17 & 6.35 & 97\\
\hline
\end{tabular}

}

\bigskip
Table 4\quad  Longwood distance comparison (alpha value: 250)
\begin{tabular}{|cc|cc|c|}
\hline
\multicolumn{2}{|c|}{Parameters}& \multicolumn{2}{c|}{Average Distances} & Ratio\\
Distance & \multicolumn{1}{c|}{Percent} & Room & \multicolumn{1}{c|}{No Room} & \%\\
\hline
\multirow{5}*{1}
& 2 & 0.78 & 0.81 & 96\\
& 5 & 1.39 & 1.70 & 82\\
& 8 & 1.91 & 1.90 & 101\\
& 11 & 2.11 & 2.21 & 96\\
& 14 & 2.00 & 2.38 & 84\\
\hline
\multirow{5}*{2}
& 2 & 1.11 & 1.12 & 99\\
& 5 & 1.81 & 1.99 & 91\\
& 8 & 2.42 & 2.36 & 102\\
& 11 & 2.71 & 3.02 & 90\\
& 14 & 3.03 & 3.21 & 94\\
\hline
\multirow{5}*{3}
& 2 & 1.44 & 1.67 & 86\\
& 5 & 2.23 & 2.65 & 84\\
& 8 & 3.20 & 3.40 & 94\\
& 11 & 3.48 & 3.67 & 95\\
& 14 & 3.68 & 4.32 & 85\\
\hline
\multirow{5}*{4}
& 2 & 2.02 & 2.04 & 99\\
& 5 & 6.63 & 6.70 & 99\\
& 8 & 4.26 & 4.58 & 93\\
& 11 & 5.77 & 6.24 & 92\\
& 14 & 5.73 & 6.24 & 92\\
\hline
\end{tabular}
\end{center}

Ratio values smaller than 100 percent mean that, on average, the vertices of the room detection maps are closer to the position of the vertex in the base map than compared to the Topology Graphs without room detection. Thus one can conclude that for those maps the vertices with room detection are more consistent, reliable and repeatable than the vertices without room detection.

From the results we can observe that the room detection graphs have clear advantages over the non-room detection graphs. This of course also depends on the maps: if there are few rooms which are also themselves subject to the noise (Fig. 12 "Longwood", Table 4), because there are many pillars and obstacles close to the room detection alpha value, then the room detection approach gives little advantages. For those maps one would first need to filter out smaller obstacles (e.g. using a "pillar detection").

Other maps with big rooms, on the other hand, profit extremely well from the room detection. This is the case with Fig. 14 (Beeson Map, Table 1), and Fig. 10 (Belgioioso Map, Table 2). Generally, higher noise distances lessen the positive effect of room detection because they increasingly effect the topology of the rooms. But the percentage of noise added has little overall effect on the ratio.

The generation of the Topology Graphs of two maps, including room detection and matching is very fast and takes less than one second for each of the maps in this paper.

\begin{center}\small

 {\interlinepenalty=10000
Table 5\quad  SDR distance comparison (alpha value: 422)\\
\begin{tabular}{|cc|cc|c|}
\hline
\multicolumn{2}{|c|}{Parameters}& \multicolumn{2}{c|}{Average Distances} & Ratio\\
Distance & \multicolumn{1}{c|}{Percent} & Room & \multicolumn{1}{c|}{No Room} & \%\\
\hline
\multirow{5}*{1}
& 2 & 1.69 & 2.77 & 61\\
& 5 & 2.96 & 3.45 & 86\\
& 8 & 2.79 & 3.46 & 81\\
& 11 & 4.19 & 4.59 & 91\\
& 14 & 3.34 & 3.80 & 88\\
\hline
\multirow{5}*{2}
& 2 & 2.11 & 2.73 & 77\\
& 5 & 3.58 & 4.08 & 88\\
& 8 & 3.15 & 3.70 & 85\\
& 11 & 4.64 & 5.08 & 91\\
& 14 & 4.73 & 4.84 & 98\\
\hline
\multirow{5}*{3}
& 2 & 2.42 & 3.58 & 68\\
& 5 & 4.25 & 4.80 & 89\\
& 8 & 5.52 & 5.84 & 94\\
& 11 & 4.51 & 5.38 & 84\\
& 14 & 4.99 & 5.55 & 90\\
\hline
\multirow{5}*{4}
& 2 & 3.85 & 4.35 & 89\\
& 5 & 5.48 & 5.96 & 92\\
& 8 & 7.10 & 7.07 & 101\\
& 11 & 6.28 & 6.47 & 97\\
& 14 & 5.97 & 6.12 & 98\\
\hline
\end{tabular}
}

\end{center}
\end{subsection}

\section{Conclusion}

In this paper we showed how we can detect rooms in 2D grid maps using the alpha shape algorithm and then use those extracted room polygons to modify the Topology Graph. The such generated Topology Graphs represent the rooms with a single vertex but keep the topological information about the connection to the adjacent vertices intact by adding new vertices at the intersection between the room polygon and the edges leading into the room.

Using artificial maps and real RoboCup Rescue maps we showed that the algorithm works well. In the final experiments we objectively evaluated the stability and repeatability of the such generated Topology Graphs by evaluating the matches of 1000 map pairs.

Using the increased stability we can integrate the Topology Graphs with room detection to the graph matching framework introduced in \cite{Schwertfeger2015_Topo_AuRo}. We will also combine it with our work on path similarity \cite{SchwertfegerPathsIAV2016}. This will then enable us to generate very robust map matches for various application scenarios.

Translation addendum: The work presented here has been used in the work on generating Area Graphs \cite{Hou2019areagraph} and subsequently using those Area Graphs to match maps \cite{Hous2019areamatching}. 

\bibliographystyle{IEEEtran}

\bibliography{ref}             

\end{document}